\documentclass[twoside,journal]{IEEEtran}
\usepackage{amsmath,amsfonts}
\usepackage{algorithmic}
\usepackage{algorithm}
\usepackage{array}
\usepackage{textcomp}
\usepackage{stfloats}
\usepackage{url}
\usepackage{verbatim}
\usepackage{graphicx}
\usepackage{cite}
\usepackage{subcaption}
\usepackage{booktabs}
\usepackage{rotating}
\usepackage{multirow}
\usepackage{url}
\usepackage[hidelinks]{hyperref}
\usepackage{multirow}
\usepackage[table]{xcolor}
\usepackage{makecell}
\begin{document}
\title{DiffReID: Discriminative Diffusion Model for Object Re-Identification}
\author{Yingquan Wang, Pingping Zhang$^{*}$,\IEEEmembership{~IEEE Member}, Dong Wang,\IEEEmembership{~IEEE Member}, Huchuan Lu,\IEEEmembership{~IEEE Fellow}
\thanks{
Yingquan~Wang, Dong~Wang and Huchuan~Lu are with the School of Information and Communication Engineering, Dalian University of Technology. (Email: yingquan\_w95@mail.dlut.edu.cn; wdice@dlut.edu.cn; lhchuan@dlut.edu.cn)

Pingping~Zhang is with the School of Future Technology, Dalian University of Technology. (Email: zhpp@dlut.edu.cn)
}}

\markboth{IEEE Transactions on Image Processing}{}
\maketitle
\begin{abstract}
As a fundamental image processing task, object Re-Identification (ReID) aims to retrieve objects across non-overlapping cameras.
Recently, with the development of deep learning, significant advancements have been made in object ReID.
However, most existing methods suffer from generalization due to the limited size and diversity of ReID datasets.
Meanwhile, current models tend to focus on extracting semantic patterns rather than learning identity-aware feature distributions.
To address these issues, we propose a novel feature learning framework named \textbf{DiffReID} for object ReID.
It leverages a discriminative diffusion model to gradually learn identity-aware distributions and generate identity-invariant features.
More specifically, with the Contrastive Language-Image Pre-training (CLIP) model, we first obtain identity-aware text features by prompt tuning.
Then, we propose a Vision-guided Noise Generator (VNG) to initialize probabilistic noises and gradually corrupt identity-aware text features.
Afterwards, we take visual features as conditions and propose a Light Weight Denoiser (LWD) to denoise the corrupted text features step-by-step for identity-aware distribution learning.
To obtain discriminative features, we further generate identity-invariant guided features from randomly sampling visual-guided noises.
Finally, we propose a Mutual Enhancement Constraint (MEC) to facilitate mutual learning between visual features and guided features to enhance the representation robustness and discrimination.
Extensive experiments on five object ReID benchmarks demonstrate that our method shows better results than most state-of-the-art methods.
The source code is available at https://github.com/AWangYQ/DiffReID.
\end{abstract}
\section{Introduction}
\begin{figure}[htbp]
    \centering
    \includegraphics[width=0.9\linewidth]{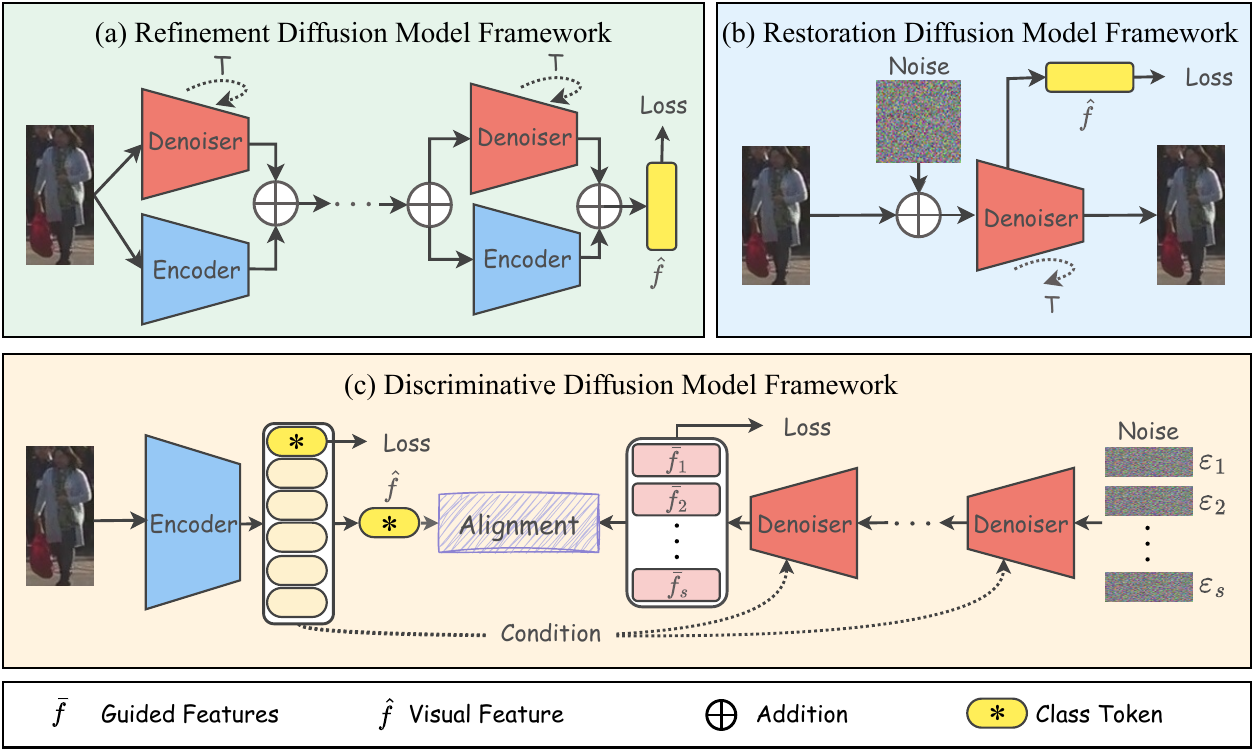}
    \caption{Different diffusion model frameworks for object ReID.
    (a) The refinement diffusion model framework unifies feature extraction and denoising within a backbone pipeline.
    (b) The restoration diffusion model framework utilizes intermediate features from diffusion models for object representation.
    (c) Our discriminative diffusion model framework generates identity-invariant guided features and facilitates mutual learning between guided features and visual features.
    }
    \label{fig:intro}
\end{figure}
\IEEEPARstart{O}{bject} Re-Identification (ReID) aims to match objects across non-overlapping cameras.
As a fundamental image processing task, it plays an important role in many real-world applications, including societal security, intelligent surveillance, mobile robotics and human-computer interaction.
The key of ReID models is to extract identity-invariant features in complex visual variations, such as illumination changes, cross-scale resolutions and different occlusions.
Although achieving great successes, most existing methods~\cite{he2021transreid,clip_reid,tfclip,wang2024other,cao2025an} suffer from generalization due to the limited size and diversity of ReID datasets.
Meanwhile, current methods tend to focus on extracting semantic patterns rather than learning identity-aware feature distributions.

Recently, diffusion models~\cite{ddpm} have gained more attention for their superior ability to model complex data distributions in many image generation tasks~\cite{cao2024survey}.
As a result, diffusion models can learn discriminative visual representations.
Thus, researchers have begun to introduce diffusion models into object ReID~\cite{wang2024denoiserep,tao2024unsupervised,niu2025synthesizing}.
Currently, two types of diffusion model-based frameworks have been proposed for object ReID.
As shown in Fig.~\ref{fig:intro}(a), one type directly unifies feature extraction and denoising within a backbone pipeline~\cite{wang2024denoiserep}.
However, the layer-wise injection of Gaussian noises may weaken the representational ability of hierarchical features.
On the other hand, as shown in Fig.~\ref{fig:intro}(b), several works~\cite{feng2024multi,mukhopadhyay2024text,kim2025revelio} have demonstrated that intermediate features of pre-trained diffusion models contain rich semantic information, making them well-suited for downstream discriminative tasks.
However, pre-trained diffusion models typically rely on complex network architectures and large timesteps, significantly increasing inference time and computational cost.

To address the above issues, we propose a novel feature learning framework named DiffReID for high-performance object ReID.
Different from previous diffusion model-based ReID methods, our framework constructs a discriminative feature distribution learning process in a compact latent space.
As shown in Fig.~\ref{fig:intro}(c), it leverages diffusion models to learn identity-aware feature distributions and generate identity-invariant guided features for object representation learning.
Specifically, we introduce three key modules: the Vision-guided Noise Generator (VNG) for noise initialization, the Light Weight Denoiser (LWD) for feature generation, and the Mutual Enhancement Constraint (MEC) for feature refinement.
They are jointly designed to support more discriminative representation learning.
Technically, inspired by the strong vision-text understanding ability of diffusion models, we first obtain identity-aware text features via prompt tuning.
As demonstrated in~\cite{li2024cliff}, noise initialization plays a crucial role in diffusion models.
Thus, we utilize VNG to take visual features as priors and sample vision-aware noises to gradually corrupt identity-aware text features.
Afterwards, we propose LWD to model the identity-aware distribution.
The LWD allows the model to predict noises and generate identity-invariant guided features.
Finally, we propose MEC to facilitate mutual learning between visual features and guided features to enhance the representation robustness and discrimination.
Experiments on five object ReID benchmarks demonstrate that our method shows better results than most state-of-the-art methods.

In summary, our main contributions are as follows:
\begin{itemize}
\item We propose a novel feature learning framework named DiffReID for high-performance object ReID.
It leverages the underlying ability of diffusion models to obtain discriminative features while maintaining efficient inference.
\item We introduce three key modules to facilitate discriminative feature extraction: a Vision-guided Noise Generator (VNG) for noise initialization, a Light Weight Denoiser (LWD) for feature generation, and a Mutual Enhancement Constraint (MEC) for feature refinement.
\item Experiments on five object ReID benchmarks demonstrate that our method achieves outstanding performance.
\end{itemize}
\section{Related Work}
\label{sec:related work}
\subsection{Diffusion Models for Representation Learning}
Diffusion models~\cite{ddpm} belong to a class of generative models.
They have emerged as mainstream generative methods due to their exceptional ability in modeling complex data distributions.
Recent studies have explored how diffusion models can be adapted for discriminative tasks~\cite{li2024cliff,han2024latent}.
In fact, current works can be broadly grouped into two categories.
One category of methods directly leverages pre-trained diffusion models (\emph{e.g.}, Stable Diffusion~\cite{stable_diffusion} and Imagen~\cite{imagen}) as feature extractors.
For example, Li~\emph{et al.}~\cite{li2023your} show that diffusion models can be leveraged to perform zero-shot classification without any additional training.
Soumik~\emph{et al.}~\cite{mukhopadhyay2024text} and Dahye~\emph{et al.}~\cite{kim2025revelio} further extract discriminative intermediate features from diffusion models for robust segmentation and classification.
While pre-trained diffusion models show great potential for discriminative tasks, their complex denoisers and large timestep requirements lead to high computational cost and long inference time.
Another category of methods focuses on designing task-specific denoising models that learn dataset-specific distributions to generate both diverse and discriminative features.
For instance, Du~\emph{et al.}~\cite{du2023protodiff} employ diffusion models for class-wise prototype feature generation, significantly improving few-shot classification performance.
Li~\emph{et al.}~\cite{li2024cliff} introduce diffusion models for continuous distribution transformation across object, image, and text latent spaces.
These methods utilize denoisers to model task-specific data distributions, effectively balancing feature generation and computational efficiency.
However, most of them naively utilize diffusion models for diverse image generation or general feature extraction.
In contrast, our method constructs a discriminative feature distribution learning process for diffusion models.
It can effectively utilize fine-grained conditions to generate more discriminative features.
\subsection{Image-based Object Re-Identification}
Early object ReID methods predominantly leverage Convolutional Neural Networks (CNNs) to extract fine-grained feature representations~\cite{sun2018beyond,wang2018learning}.
However, the inherent locality of CNNs limits their ability to capture long-range dependencies.
To this end, recent studies have integrated attention mechanisms~\cite{vaswani2017attention} for robust object ReID.
For example, Zhang~\emph{et al.}~\cite{zhang2020relation} construct a global pixel attention map to obtain more robust object representations.
He~\emph{et al.}~\cite{he2021transreid} introduce a Transformer-based backbone to capture global dependencies, achieving impressive performance in object ReID.
Afterwards, many Transformer-based ReID models~\cite{zhang2021hat,li2023dc,lu2023learning,yan2023learning,wang2025unity} are proposed.
However, image-only methods easily overfit to salient regions while neglecting essential semantic information.
Therefore, many researchers~\cite{clip_reid,tfclip,yu2025climb} have incorporated large vision-language models into object ReID.
Besides, recent studies have also explored the model generalization.
For example, Li~\emph{et al.}~\cite{li2025breaking} investigate domain generalization object ReID by leveraging cross-camera unpaired samples.
Zhang~\emph{et al.}~\cite{zhang2025weakly} propose heterogeneous expert collaborative learning for weakly supervised visible-infrared person ReID.
Zhang~\emph{et al.}~\cite{zhang2025dual} utilize a dual-granularity cross-modal identity association for text-based object ReID.
Li~\emph{et al.}~\cite{li2024catalyst} improve clustering-based unsupervised object ReID via feature calibration.
However, object ReID remains constrained by the limited availability of large-scale training datasets.
To mitigate this issue, recent works~\cite{siddiqui2025dlcr,niu2025synthesizing} leverage diffusion models to enrich training data.
For example, Si~\emph{et al.}~\cite{siddiqui2025dlcr} introduce a diffusion model-based inpainting method to generate clothing-change samples.
Similarly, Niu~\emph{et al.}~\cite{niu2025synthesizing} leverage diffusion models to construct large-scale person ReID datasets.
However, simple image generation inevitably causes the loss of identity information.
To this end, some researchers leverage diffusion models for both feature extraction and image generation.
For example, Wang~\emph{et al.}~\cite{wang2024denoiserep} integrate feature extraction and denoising into a single image encoder.
Additionally, Jia~\emph{et al.}~\cite{jia2024psdiff} utilize diffusion models to learn the distribution of bounding boxes and ReID features, conditioned on the provided visual features.
Unlike existing works that solely utilize generated features for objects, we explore the mutual learning between generated features and real visual features for more robust and discriminative representations.
\begin{figure*}[htbp]
\centering
\includegraphics[width=0.9\linewidth]{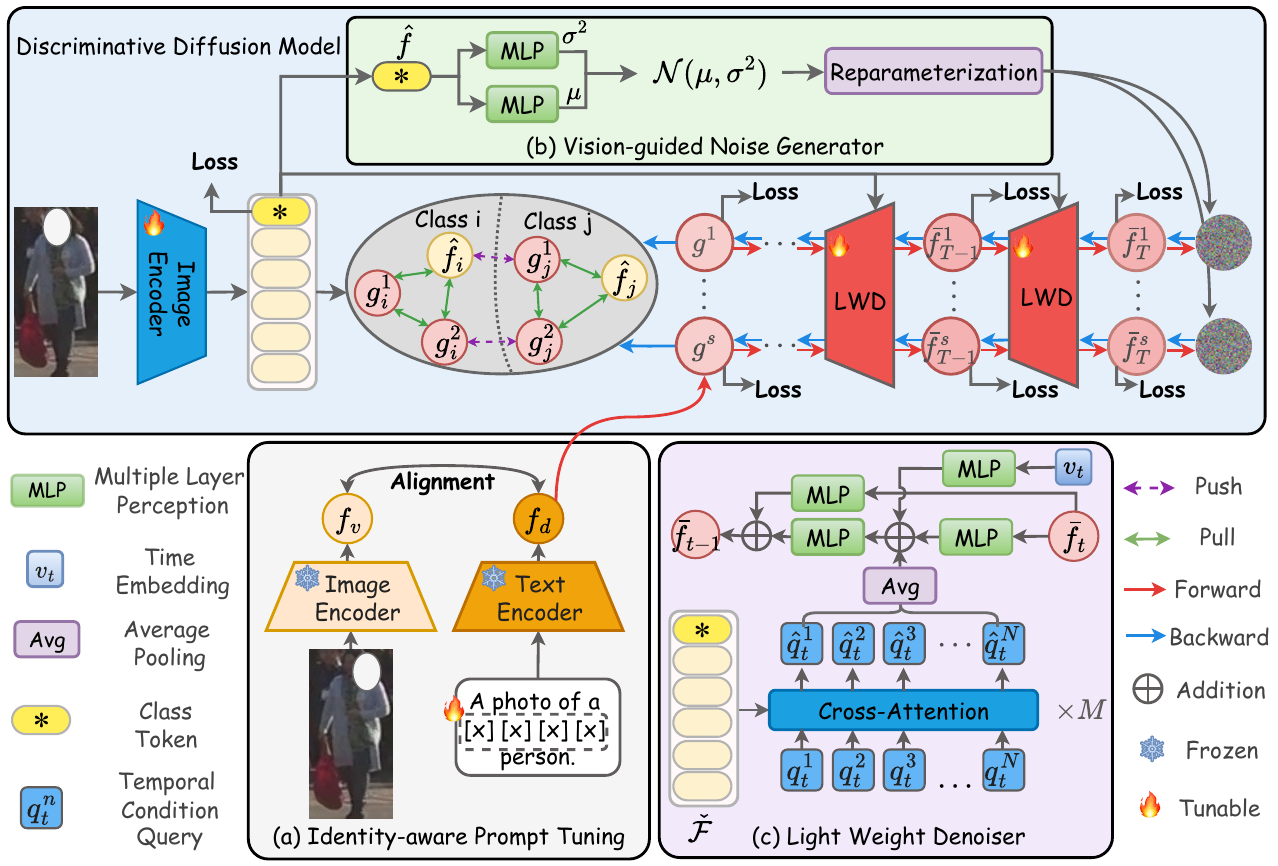}
\caption{Illustration of our proposed framework. With the CLIP model, the framework first generates identity-aware text features by prompt tuning.
Then, the Vision-guided Noise Generator (VNG) is used to initialize probabilistic noises and gradually corrupt identity-aware text features.
Afterwards, the Light Weight Denoiser (LWD) is used to denoise the corrupted text features for identity-aware distribution learning.
To train the framework, the Mutual Enhancement Constraint (MEC) is performed to facilitate mutual learning between visual features and guided features.}
\label{fig:framework}
\end{figure*}
\section{Our Proposed Method}
As illustrated in Fig.~\ref{fig:framework}, our framework leverages a discriminative diffusion model to gradually learn identity-aware distributions and generate identity-invariant features.
More specifically, with the CLIP model~\cite{clip}, we first obtain identity-aware text features by prompt tuning.
Then, we propose a Vision-guided Noise Generator (VNG) to initialize probabilistic noises and gradually corrupt identity-aware text features.
Afterwards, we propose a Light Weight Denoiser (LWD) to denoise the corrupted text features for identity-aware distribution learning.
Finally, we propose a Mutual Enhancement Constraint (MEC) to facilitate mutual learning between visual features and guided features to enhance the representation robustness and discrimination.
We describe the above modules in the following sections.
\subsection{Overview of Our Framework}
The overall framework adopts a two-stage training procedure.
In the first stage, we aim to obtain identity-aware text features by prompt tuning.
As shown in Fig.~\ref{fig:framework}(a), with the CLIP model, we freeze the pre-trained text encoder and image encoder, only optimizing prompt tokens.
Inspired by~\cite{clip_reid}, these prompt tokens are passed through the text encoder with the template ``A photo of a [x] [x] [x] [x] person/vehicle''.
To obtain identity-aware text features, we employ the following bidirectional contrastive losses:
\begin{equation}
L_{t2i}(y) = -\frac{1}{|P(y)|}\sum_{p\in P(y)}\text{log}\frac{exp(f^p_v,f^{y}_d)}{\sum^B_{b=1}exp(f^a_v, f^{y}_d)},
\end{equation}
\begin{equation}
L_{i2t}(y) = -\frac{1}{|P(y)|}\sum_{p\in P(y)}\text{log}\frac{exp(f^p_d,f^{y}_v)}{\sum^B_{b=1}exp(f^a_d, f^{y}_{v})},
\end{equation}
where $P(y)$ is the image set of the same identity $y$.
$B$ is the batch size.
$f_d$ and $f_v$ are text and visual features, respectively.

In the second stage, we feed person images into the learnable image encoder to extract visual tokens $\mathcal{F}\in\mathbb{R}^{N\times D}$.
They are sent to the VNG and LWD for noise initialization and feature generation, respectively.
It involves two key processes: a forward noising process and a backward denoising process.
Unlike most diffusion models~\cite{jia2024psdiff}, we integrate textual semantics into the denoising process.
Specifically, as shown in the {red} line of Fig.~\ref{fig:framework}, the forward noising process gradually adds visual-guided noises $\varepsilon$ to the text feature $f_d$ over $T$ timesteps.
It should be noted that VNG generates visual-guided noises based on visual priors.
At each timestep, visual-guided noises are injected with a timestep-dependent variance $\alpha_t$:
\begin{equation}
q(\bar{f}_t|f_d) = \mathcal{N}(\bar{f}_t; \sqrt{\bar{\alpha}_t}f_d, (1 - \bar{\alpha}_t){I}),
\end{equation}
\begin{equation}
\bar{f}_t = \sqrt{\bar{\alpha}_t}f_d + \sqrt{1 - \bar{\alpha}_t}\varepsilon,
\end{equation}
where $\bar{\alpha}_t = \prod_{i=1}^t \alpha_i$.
This process ultimately transforms $f_d$ into the final state $\bar{f}_T$, which closely approximates its visual-aware Gaussian distribution $\mathcal{N}(\mu, \sigma^2)$.
Noting that, $\mu$ and $\sigma^2$ are obtained by the VNG.
As shown in the blue line of Fig.~\ref{fig:framework}, we then treat the visual feature as a condition to the backward denoising process.
It aims at gradually denoising $p(\bar{f}_T)$ to $q(\bar{f}_0)$ as follows:
\begin{equation}
    p_{\theta}(\bar{f}_{t-1}|\bar{f}_t, \hat{c}_t) = \mathcal{N}\Big(\bar{f}_{t-1}; \mu_{\theta}(\bar{f}_t, \hat{c}_t, t),\begin{matrix}\Sigma_{\theta}(\bar{f}_t, \hat{c}_t, t)\end{matrix}\Big),
\end{equation}
where $p_{\theta}(\bar{f}_{t-1}\mid\bar{f}_t,\hat{c}_t)$ is the backward denoising distribution.
$f_d \sim q(\bar{f}_0)$, $\theta$ denotes learnable parameters, and $\hat{c}_t$ is a timestep-specific condition.
To obtain guided features, our framework first generates an initial noise state $\bar{f}_T$, and then progressively samples $\bar{f}_{t-1}$ from $p_{\theta}(\bar{f}_{t-1}\mid\bar{f}_t,\hat{c}_t)$ at each denoising timestep $h\in\{1,\ldots,T\}$.
After all denoising timesteps, the final denoised state $\bar{f}_0$ is taken as the guided feature $g$.
Here, $t$ and $h$ are the numbers of noising and denoising steps, respectively.
They satisfy $T=t+h$.
Note that, the above diffusion process is performed in the latent feature space rather than the image space.
It avoids reconstructing high-dimensional pixel-level details and provides a more compact semantic guidance for identity-aware distribution learning.
Moreover, we perform the diffusion process on identity-aware text features rather than visual features.
In fact, visual features may contain appearance-biased cues, such as clothing colors, bags, or backgrounds, while identity-aware text features provide a compact semantic distribution aligned with identity labels.
Therefore, with visual features as conditions and identity-aware text features as diffusion targets, our framework can generate identity-aware guided features that preserve visual discrimination while benefiting from semantic guidance.
\subsection{Vision-guided Noise Generator}
\label{sec:4.2}
As demonstrated in~\cite{li2024cliff}, the noise initialization significantly affects the generation process of diffusion models.
In our framework, the initialized noise is used to corrupt identity-aware text features and determines the starting state of the denoising process.
Therefore, the quality of initial noises directly influences whether the guided features can preserve identity-related information.
However, most existing methods~\cite{ddpm,du2023protodiff} sample noises directly from a standard Gaussian distribution.
Such randomly sampled noises are sample-agnostic and contain no visual prior related to the input image.
They may introduce perturbations inconsistent with the object identity and degrade the discrimination of generated features.
To address this issue, we introduce the Vision-guided Noise Generator (VNG) to estimate the mean and variance of noise distributions, as shown in Fig.~\ref{fig:framework}(b).
Instead of using a fixed Gaussian prior, our VNG estimates the mean and variance of the noise distribution $p_{\phi}(\bar{f}_T|\hat{f})$ based on the visual feature $\hat{f}$.
The noise is then sampled from this learned distribution, making it more adaptive for feature generation.
In particular, we use two fully connected layers to transform the visual feature $\hat{f}$ into the mean $\mu$ and variance $\sigma^{2}$:
\begin{equation}
\mu = W_{\mu} \hat{f} + b_{\mu},
\end{equation}
\begin{equation}
\sigma^{2} = W_{\sigma} \hat{f} + b_{\sigma},
\end{equation}
where $W_{\mu}$, $W_{\sigma}$, $b_{\mu}$ and $b_{\sigma}$ are learnable parameters.
Then, we sample initial visual-guided noises from the distribution $p_{\phi}(\bar{f}_T|\hat{f})$ using the reparameterization trick~\cite{kingma2013auto}:
\begin{equation}
\varepsilon = \mu + \sigma \otimes \epsilon,
\end{equation}
where $\epsilon\sim\mathcal{N}(0, I)$ and $\otimes$ denotes the element-wise multiplication.
To align with the distribution assumption in diffusion models~\cite{ddpm}, we impose a Kullback-Leibler (KL) divergence regularization to constrain $p_{\phi}(\bar{f}_T|\hat{f})$ toward a standard Gaussian distribution.
With the proposed VNG, our framework can take visual features as priors and gradually corrupt identity-aware text features.
\subsection{Light Weight Denoiser}
\label{sec:4.3}
Many works~\cite{mukhopadhyay2024text,kim2025revelio,han2025on} have demonstrated that diffusion models can learn comprehensive data distributions, improving the robustness of generated features.
However, applying complex diffusion models to discriminative tasks presents two major challenges: (a) the inconsistency between training and inference processes; (b) the high computational cost during inference.
To address these issues, we propose the Light Weight Denoiser (LWD) to denoise identity-aware text features, while maintaining computational efficiency.
As shown in Fig.~\ref{fig:framework}(c), LWD utilizes several Multilayer Perceptron (MLP) to transform identity-aware text features instead of complex networks.
To enhance the modeling capacity, we stack $M$ denoisers for LWD.
Note that, LWD is lightweight because our diffusion process is performed in a compact latent feature space for identity-aware distribution learning~\cite{du2023protodiff,li2024cliff}.
Unlike image-based diffusion models that require complex denoising networks and long denoising trajectories, LWD models the transformation between semantic feature representations.
Therefore, an MLP-based lightweight denoiser with a small number of timesteps is sufficient to generate discriminative features while maintaining computational efficiency.
To obtain the condition of LWD, we first aggregate the intermediate features from the image encoder and then pass them through a linear layer to obtain the condition representation $\check{\mathcal{F}}$,
\begin{equation}
\hat{\mathcal{F}} = \sum_{l}f_{l},
\end{equation}
\begin{equation}
\check{\mathcal{F}} = W_1 \hat{\mathcal{F}},
\end{equation}
where $f_{l}$ represents the output from the $l$-th Transformer layer and $W_1$ is the learnable parameter of a linear layer.

During the denoising process, the features change progressively across timesteps, and different stages require different types of visual guidance.
At early timesteps, the corrupted features contain stronger noise, so the denoiser mainly needs coarse and global contextual information to recover the overall identity-related structure.
At later timesteps, the features are gradually refined, and the denoiser requires more fine-grained and identity-discriminative cues to further improve feature discrimination.
Therefore, using the same condition for all timesteps may limit the ability of LWD to adapt to the evolving feature state.
To address this issue, we introduce a set of learnable Temporal Condition Queries (TCQ) to obtain timestep-specific conditions for LWD.
Specifically, at each timestep $t$, we initialize a set of learnable queries $\mathcal{Q} = \{q^1_t, q^2_t,\dots, q^N_t\}$, where $N$ denotes the number of queries per timestep.
Inspired by~\cite{huang2025deim}, we introduce the noise $\bar{f}_t$ as a prior into the corresponding condition queries to stabilize training and accelerate convergence:
\begin{equation}
\check{q}^n_t = q^n_t+W_2\bar{f}_t,
\end{equation}
where $t\in \{1, 2\cdots, T\}$, $n\in \{1, 2\cdots, N\}$ and $W_2$ is the learnable parameter.
Finally, these temporal queries $\check{q}^n_t$ interact with the condition representation  $\check{\mathcal{F}}$ through a cross-attention to obtain the condition features at each timestep:
\begin{equation}
\hat{q}^n_t = \text{Softmax} \left( \frac{\check{q}^n_t W_q (\check{\mathcal{F}} W_k)^T}{\sqrt{D}} \right) (\check{\mathcal{F}} W_v),
\end{equation}
where $W_q, W_k$ and $W_v$ are learnable parameters.
$D$ is the feature dimension for scaling.
As shown in Fig.~\ref{fig:framework}(c), we obtain the timestep-specific condition $\hat{c}_t$ by averaging the condition features as follows:
\begin{equation}
\hat{c}_t = \text{Avg}(\hat{q}^1_t, \hat{q}^2_t, \cdots, \hat{q}^N_t).
\end{equation}
Then, the noised feature $\bar{f}_t$, the time embedding $v_t$, and the timestep-specific condition $\hat{c}_t$ are jointly processed to denoise the corrupted identity-aware text features:
\begin{equation}
\label{eq:9}
    \bar{f}_{t-1} =W_3(W_4\bar{f}_t+W_5v_t+\hat{c}_t) + W_6\bar{f}_t,
\end{equation}
where $W_3$, $W_4$, $W_5$ and $W_6$ are learnable parameters.
With LWD, we can take visual features as conditions and denoise the corrupted text features step-by-step for identity-aware distribution learning.
\subsection{Mutual Enhancement Constraint}
\label{sec:4.5}
Technically, most of diffusion model-based feature learning methods~\cite{wang2024denoiserep,du2023protodiff} directly adopt generated features as the final representations.
However, they overlook the complementary between features extracted by general visual encoders and those generated by diffusion models.
In fact, features extracted by general visual encoders capture discriminative information but tend to overfit to salient appearance patterns~\cite{geirhos2020shortcut}.
In contrast, generated features of diffusion models are encouraged to follow the identity-aware feature distribution~\cite{han2025on}, making them more robust and semantic.
To better integrate the advantages of both features, we introduce the Mutual Enhancement Constraint (MEC), which encourages visual features and generated features to learn from each other.
Specifically, generated features provide identity-aware semantic cues to improve the generalization of visual features. %
Visual features provide image-specific evidence to constrain the generated features and suppress unreliable semantic deviations.

In each iteration, the diffusion model first generates guided features that follow the identity-aware distribution.
Then, these guided features and visual features mutually refine each other.
Specifically, for each input image $I_b$, we obtain a feature set as follows:
\begin{equation}
\mathcal{V}_b= \{\hat{f}_b, g^1_b, \cdots, g^S_b\},
\end{equation}
where $b\in\{1,2,\cdots,B\}$, $\hat{f}$ is the visual feature and $g^s$ is the guided feature.
$S$ is the total number of guided features.
Then, we can construct the feature set within each batch:
\begin{equation}
    \mathcal{V} = \{\mathcal{V}_1, \mathcal{V}_2 \cdots, \mathcal{V}_B\}.
\end{equation}
To achieve mutual learning, we introduce the MEC:
\begin{equation}
L_{\text{mec}} = \sum_{\substack{a,p,n \\ y_a = y_p \neq y_n}} [D_{a,p} - D_{a,n} + m_1]_+,
\end{equation}
where $D_{a,p}$ and $D_{a,n}$ are the distances between the anchor–positive and anchor–negative pairs within $\mathcal{V}$, respectively.
$m_1$ is the margin parameter.
As observed, MEC combines the visual and guided features into a joint feature set and constructs cross-feature triplets.
The triplet loss pulls each guided feature closer to the same-identity visual feature and pushes it away from different identity features.
Thus, the visual feature acts as a stable identity reference that reduces stochastic shifts of the guided feature toward incorrect identities.
Meanwhile, the guided feature introduces distributional variations that encourage the visual feature to retain identity-consistent cues.
Through this bidirectional interaction, the advantages of visual and guided features are jointly amplified, enabling the framework to learn more discriminative representations.
\subsection{Training and Inference}
\label{sec:4.6}
\textbf{Model Training.}
In the first stage, we follow previous works~\cite{clip_reid,wang2025makes} and apply the bidirectional contrastive losses for prompt tuning:
\begin{equation}
L_{\text{stage1}} = L_{t2i} + L_{i2t}.
\end{equation}

In the second stage, we employ multiple losses to supervise the framework.
To optimize the visual feature $\hat{f}$, we adopt the following combined loss:
\begin{equation}
L_{\text{vis}} = \gamma_1 L_{ce} + \gamma_2 L_{tri} + L_{t2i},
\end{equation}
where $L_{ce}$ is the identity loss.
$L_{tri}$ is the triplet loss.
$\gamma_1$ and $\gamma_2$ are hyper-parameters to balance the loss terms.

As for the diffusion procedure, we supervise the noise prediction at each timestep by the following loss:
\begin{equation}
L_{\text{diffusion}} = \frac{1}{T}\sum^{T}_{t=1}||\bar{f}_t - f_d||_2.
\end{equation}

As for guided features, we further ensure that they have the same identity with the following loss:
\begin{equation}
L_{\text{dis}} = \frac{1}{S}\sum^{S}_{s=1}[L_{tri}(g^s)+L_{ce}(g^s)].
\end{equation}

To align with the distribution assumption, the Kullback-Leibler (KL) divergence loss is reformulated as:
\begin{equation}
L_{kl}=D_{\text{kl}}\big(p_{\phi}(\bar{f}_T|\hat{f}) || \mathcal{N}(0, I)\big).
\end{equation}
Thus, the overall loss for the second stage is formulated as:
\begin{equation}
L_{\text{stage2}} = L_{\text{vis}} + L_{\text{diffusion}} + L_{\text{mec}} + L_{\text{dis}} + L_{\text{kl}}.
\end{equation}

\textbf{Model Inference.}
We extract the visual feature $\hat{f}$ and use LWD to generate guided features from visual-guided noises.
The final feature $f_r$ is formed by averaging the guided features:
\begin{equation}
f_r = \text{Avg}(g^1, g^2, \dots, g^S).
\end{equation}
Since each guided feature is sampled from the learned identity-aware distribution, a single guided feature may only reflect one possible representation of this distribution.
Thus, we average multiple guided features to obtain a more comprehensive and stable representation.

\begin{table*}[htbp]
    \caption{Comparison of state-of-the-art methods on three person ReID datasets and one vehicle ReID dataset. The best and second results are marked in \textbf{bold} and \underline{underlined}, respectively.}
    \centering
    \resizebox{0.84\textwidth}{!}{
    \begin{tabular}{c|cccccc|c|cc}
    \hline
    \multirow{2}{*}{Method}&
    \multicolumn{2}{c}{Market1501}&
    \multicolumn{2}{c}{MSMT17}&
    \multicolumn{2}{c|}{DukeMTMC}&
    \multirow{2}{*}{Method}&
    \multicolumn{2}{c}{VeRi-776}\\
    &             mAP         &Rank-1         & mAP         & Rank-1         & mAP         & Rank-1 & &mAP &Rank-1\\
    \hline
    Nformer~\cite{wang2022nformer}           &{{91.1}}        &94.7          &59.8         &77.3           &{83.5}         &89.4 &CFVMNet~\cite{sun2020cfvmnet}     &77.1        &95.3\\
    CMT~\cite{yan2023learning}       &87.7        &95.4          &62.6         &83.1           &80.0         &90.1 &DCAL~\cite{zhu2022dual}         &80.2        &{96.9}\\
    TransReID~\cite{he2021transreid}       &88.9        &95.2          &67.4         &85.3           &82.0         &{90.7} &GLTrans~\cite{wang2024other} &82.9 &\underline{97.5}\\
    GLTrans~\cite{wang2024other}            &90.0        &95.6          &69.0         &85.8           &82.4         &90.7 &HPGN~\cite{HPGN} &80.2 &96.7\\
    HAT~\cite{zhang2021hat}              &89.8        &{95.8}          &61.2         &82.3           &81.4         &90.4 &PGAN~\cite{zhang2020part}             &79.3        &96.5\\
    ADSO~\cite{zhang2021coarse}     &87.7   &94.8   &--  &--  &74.9  &87.4 &PVEN~\cite{meng2020parsing}           &79.5        &95.6\\
    PFD~\cite{wang2022pose}            &89.6        &95.5          &65.1         &82.7           &{82.2}         &90.6 &SAVER~\cite{khorramshahi2020devil}           &79.6        &96.4\\
    DCAL~\cite{zhu2022dual}            &87.5        &94.7          &64.0         &83.1           &80.1         &89.0 &SOFCT~\cite{SOFCT}  &80.7  &96.6\\
    SAP~\cite{jia2023semi}             &{90.5}        &{{96.0}}          &67.8         &85.7           &--     &--    &GLAMOR~\cite{suprem2020looking}        &80.3        &96.5\\
    DC-Former~\cite{li2023dc}      &90.4        &{{96.0}}          &{68.8}         &{86.2}           &--     &-- &MPC~\cite{li2021exploiting}    &{80.9}  &96.2\\
    RGANet~\cite{he2023region}  &89.8 &95.5 &72.3 &88.1 &--   &-- &MsKAT~\cite{li2022mskat}     &{82.0}  &{{97.1}}\\
    PHA~\cite{zhang2023pha} &90.2 &{\underline{96.1}} &68.9 &86.1 &-- &--     &TransReID~\cite{he2021transreid}  &{82.0}       &{{97.1}}\\
    PCL-CLIP~\cite{li2023prototypical}  &91.4  &95.9  &76.1  &89.8 &-- &-- &PCL-CLIP~\cite{li2023prototypical} &82.5  &97.1\\
    CLIP-ReID~\cite{clip_reid} &{90.5} &95.4 &{75.8} &{89.7} &83.1 &{90.8}   &Vehicle-Diff~\cite{Vehicle-Diff} &83.8 &\textbf{97.7}\\
    TF-CLIP~\cite{tfclip} &{90.4} &95.7 &{73.9} &{88.5} &-- &--    &MDPDTrans~\cite{MDPDTrans}  &83.7  &\textbf{97.7}\\
    DenoiseRep~\cite{wang2024denoiserep} &{{91.1}} &{{95.8}}  &{{76.3}}  &{\underline{90.6}}  &{\underline{83.7}}  &{\underline{91.6}}     &ADPRP-Net~\cite{ADPRP-Net}     &{{82.8}}  &95.6\\
    CLIMB-ReID~\cite{yu2025climb} &\textbf{92.6} &\textbf{96.8} &\underline{77.8} &{90.5} &-- &--    &CLIP-ReID~\cite{clip_reid}    &{\underline{84.5}}  &{{97.3}}\\
    \hline
    DiffReID (Ours)          &{\underline{91.5}}        &
    {\underline{96.1}}         &{\textbf{79.0}}       &{\textbf{91.2}}         &{\textbf{85.6}}         &{\textbf{92.4}} &DiffReID (Ours) &{\textbf{84.7}}         &{{97.4}} \\
    \hline
    \end{tabular}}
    \label{table:image-based sota}
\end{table*}
\section{Experiments}
\subsection{Datasets and Evaluation Metrics}
To comprehensively evaluate the effectiveness of our proposed framework, we conduct experiments on five large-scale object ReID datasets, \emph{i.e.}, Market1501~\cite{market1501}, DukeMTMC~\cite{Duke}, MSMT17~\cite{MSMT17}, CUHK03~\cite{cuhk03} and VeRi-776~\cite{VeRi}.
The details of these datasets can be found in corresponding references.
To assess the generalization ability, we employ two domain generalization settings~\cite{Ni2022meta}.
The first setting involves training on one dataset and testing on another unseen dataset.
The second setting utilizes the training sets of multiple datasets for training and evaluates on the test set of an unseen dataset.
Following previous works~\cite{sun2018beyond,he2021transreid}, we use mean Average Precision (mAP) and Cumulative Matching Characteristics (CMC) at Rank-1 as our evaluation metrics.
\subsection{Implementation Details}
We implement our framework using the PyTorch toolbox. All experiments are conducted on a single NVIDIA RTX 4090 GPU with 24GB memory.
We adopt the pre-trained CLIP~\cite{clip} as the feature extraction backbone for both images and texts.
Person images are resized to $256 \times 128$ and vehicle images are resized to $256 \times 256$.
In the first stage, we employ the Adam optimizer~\cite{kinga2015method} with an initial learning rate of $3.5 \times 10^{-4}$ and cosine decay.
Training is performed with a batch size of 64, without augmentation, only optimizing the learnable prompt tokens.
In the second stage, a mini-batch of 64 images is sampled, which contains 16 identities and each with 4 images.
Data augmentation consists of random cropping, horizontal flipping, and random erasing~\cite{randomerasing}.
We train the model for 60 epochs using Adam, setting the initial learning rate to $5 \times 10^{-6}$ for the image encoder and $5 \times 10^{-4}$ for the discriminative diffusion model.
The model is warmed up for 10 epochs, with the learning rate reduced by a factor of 0.1 at the 30th and 50th epochs.
The hyper-parameters $\gamma_1$, $\gamma_2$, $m_1$, $T$ and $M$ are set to 0.25, 1.0, 0.3, 10 and 3, respectively.
\begin{table*}
\small
    \caption{Comparison of state-of-the-art methods on domain generalization settings. The best and second results are marked in \textbf{bold} and \underline{underlined}, respectively. The superscript * indicates that the model resizes the input images to 384$\times$128.}
    \centering
    \resizebox{0.8\textwidth}{!}{
    \begin{tabular}{l|c|cc|cc|cc|cc}
    \hline
    \multirow{2}{*}{Method}&
    \multirow{2}{*}{Training}&
    \multicolumn{2}{c|}{Market1501}&
    \multicolumn{2}{c|}{MSMT17}&
    \multicolumn{2}{c|}{CUHK03-NP}&
    \multicolumn{2}{c}{DukeMTMC}
    \\
    &       &      mAP         &R1         & mAP         & R1         & mAP         & R1  & mAP         & R1\\
    \hline
    QAConv~\cite{liao2020interpretable}   &\multirow{8}{*}{Market1501}         &--        &--          &7.0         &22.6           &8.6         &9.9  &33.6  &54.4  \\
    TransMatcher~\cite{liao2021transmatcher}      &     & --        &--          &18.4         &47.3           &21.4         &22.2 &--  &--  \\
    QAConv+Gs~\cite{Liao2022Graph}&     &--  &--  &17.2  &45.9  &18.1  &19.1 &--  &--  \\
    MDA~\cite{Ni2022meta} &              &--        &--          &11.8         &33.5           &--         &-- &34.4  &56.7  \\
    PAT~\cite{Ni2023Part} &           &--        &--          &18.2         &42.8           &{26.0}        &25.4 &{{48.9}} &{{67.9}}\\
    OGNorm$^{*}$~\cite{chen2024multi} &    &-- &--  &{{19.9}}  &{\underline{49.7}} &{24.9}  &{{26.6}}  &--  &--\\
    CLIP-ReID~\cite{clip_reid}  & &--  &--  &{\underline{23.0}}  &{{48.9}} &{\underline{38.5}} &{\underline{39.9}}  &{\underline{51.7}} &{\underline{69.6}} \\
    DiffReID (Ours)   &      & --  &--  &\textbf{24.6}  &\textbf{51.5}  &\textbf{41.2}  &\textbf{43.5} &\textbf{51.9} &\textbf{71.8}\\
    \hline
    QAConv~\cite{liao2020interpretable}   &\multirow{8}{*}{MSMT17}         &43.1        &72.6          &--         &--           &22.6         &25.3 &{{53.4}}  &{\textbf{72.2}}  \\
    TransMatcher~\cite{liao2021transmatcher}    &  &52.0        &{{80.1}}               &--         &--           &22.5         &23.7  &--  &--  \\
    QAConv+Gs~\cite{Liao2022Graph} &     &49.5   &79.1   &-- &--  &20.6  &20.9  &--  &--  \\
    MDA~\cite{Ni2022meta} &           &53.0        &{79.7}          &--         &--           &--         &--  &52.4  &71.7  \\
    PAT~\cite{Ni2023Part} &           &47.3        &72.2          &--         &--          &25.1        &24.2  &--  &--  \\
    OGNorm$^{*}$~\cite{chen2024multi} &  &{\textbf{54.5}}  &{\textbf{83.0}}  &-- &--  &{{28.5}}  &{{31.0}}  &--  &--\\
    CLIP-ReID~\cite{clip_reid}  & &{{51.5}}  &{76.2}  &--  &--  &{\underline{38.9}} &{\underline{40.2}}  &{\underline{58.0}}  &{\underline{74.9}} \\
    DiffReID (Ours)   &      &\underline{54.2}  &\underline{82.5}  &-- &-- &\textbf{40.7} &\textbf{42.7}  &\textbf{60.3} &\textbf{76.0}\\
    \hline
    QAConv~\cite{liao2020interpretable}      &\multirow{7}{*}{Multi-source}        &39.5        &68.6          &10.0         &29.9           &19.2         &22.9 &43.4  &64.9  \\
    RaMoE~\cite{dai2021generalizable}    &  &56.5  &82.0  &13.5  &34.1  &{{35.5}} &{{36.6}}  &{\underline{56.9}} &{\underline{73.6}}\\
    $\mathrm{M}^{3}\mathrm{L}$~\cite{zhao2021learning} &           &50.2        &75.9          &14.7         &36.9           &32.1       &33.1 &51.1  &69.2  \\
    CINorm~\cite{chen2023cluster} & &{{57.8}}       &{{82.3}} &21.1  &{{49.7}}  &31.1  &30.3  &52.4  &71.3\\
    PAT~\cite{Ni2023Part}  & & 51.7 &75.2 &{{21.6}} &45.6 &31.5 &31.1 &{{56.5}} &{{71.8}}\\
    OGNorm$^{*}$~\cite{chen2024multi} &  &{\textbf{65.2}}  &{\textbf{87.1}}  &{\underline{25.9}}  &{\textbf{57.7}}  &{\underline{40.3}}  &{\underline{44.0}}  &--  &--\\
    DiffReID (Ours)    &      &{\underline{61.6}}        &{\underline{82.3}}         &{\textbf{26.8}}      &{\underline{52.6}}         &{\textbf{50.7}}         &{\textbf{51.4}} &{\textbf{61.5}}  &{\textbf{76.8}}\\
    \hline
    \end{tabular}}
    \label{table:DG sota}
\end{table*}
\subsection{Comparison with State-of-the-Arts Methods}
In Tab.~\ref{table:image-based sota} and Tab.~\ref{table:DG sota}, our method is compared with other state-of-the-art methods on five benchmarks under both single domain and domain generalization settings.
Note that no post-processing techniques are employed in these experiments.

\textbf{Single Domain Comparison.}
As shown in Tab.~\ref{table:image-based sota}, our method achieves state-of-the-art results on Market1501, MSMT17, DukeMTMC and VeRi-776.
More remarkably, our method achieves $79.0\%$ in mAP on MSMT17, outperforming most of the compared methods, \emph{e.g.,} DenoiseRep~\cite{wang2024denoiserep} and CLIP-ReID~\cite{clip_reid} by $2.7\%$ and $3.2\%$, respectively.
In fact, DenoiseRep~\cite{wang2024denoiserep} also introduces diffusion models for person ReID.
However, its complex denoising process and long diffusion steps may hinder model convergence, ultimately limiting performance improvements.
CLIP-ReID~\cite{clip_reid} introduces the large-scale vision-language models into object ReID.
However, due to the lack of class-wise descriptions, CLIP-ReID tends to focus on semantic patterns rather than identity-aware feature distributions.
Unlike these methods, our method introduces a diffusion model to learn identity-aware feature distributions and further generate diverse guided features, leading to more generalized and robust representations.

\textbf{Domain Generalization Comparison.}
As shown in Tab.~\ref{table:DG sota}, we evaluate the generalization of different methods in different domain generalization settings.
The results show that our method achieves competitive performance compared with other domain generalization methods.
Especially, our method achieves the best performance in the setting of Market1501+DukeMTMC+MSMT17$\rightarrow$CUHK03, reaching 50.7\% mAP, which surpasses OGNorm by 10.4\%.
These results indicate that our method effectively learns identity-aware feature distributions and prevents the model from overfitting to semantic regions.
\subsection{Ablation Studies}
In this section, we perform ablation studies on MSMT17 and DukeMTMC datasets to investigate the effect of key components and hyper-parameters.
\begin{table}[hbtp]
\centering
\caption{Ablation results with different components.}
\resizebox{0.46\textwidth}{!}{
\begin{tabular}{c|cccc|cccc}
\hline
&\multirow{2}{*}{Baseline} &\multirow{2}{*}{LWD}  &\multirow{2}{*}{VNG} &\multirow{2}{*}{MEC} &\multicolumn{2}{c}{MSMT17} &\multicolumn{2}{c}{DukeMTMC} \\
&  &  &  &  &mAP &R1 &mAP &R1 \\
\hline
(a)&\checkmark &$\times$ &$\times$ &$\times$   &75.4  &89.3 &82.8  &90.7 \\
(b)&\checkmark &\checkmark &$\times$ &$\times$   &76.6  &90.4 &83.9  &91.1 \\
(c)&\checkmark &\checkmark &\checkmark &$\times$     &77.7  &90.8 &84.5 &91.7\\
(d)&\checkmark &\checkmark &$\times$ &\checkmark     &77.2  &90.9 &84.6  &92.0\\
(e)&\checkmark &\checkmark &$\checkmark$ &$\checkmark$      &79.0  &91.2 &85.6  &92.4\\
\hline
\end{tabular}
}
\label{table:ablation study}
\end{table}

\textbf{Effect of Key Components.}
Tab.~\ref{table:ablation study} presents the ablation results with key components.
The results show that using the LWD achieves performance gains on the MSMT17 and DukeMTMC, consistently.
Furthermore, with VNG, the model achieves an additional improvement of $1.1\%$ mAP on MSMT17 and $0.6\%$ mAP on DukeMTMC.
It confirms the effectiveness of adaptive noise initialization in enhancing feature discrimination.
With MEC, the model further improves mAP by 0.6\% on MSMT17 and 0.7\% on DukeMTMC.
These results demonstrate that mutually learning visual features and guided features allows them to complement each other.
Finally, incorporating all components achieves the best results: $79.0\%$ mAP on MSMT17 and $85.6\%$ mAP on DukeMTMC.
The consistent improvements across all configurations validate the effectiveness of our proposed components.
\begin{table}[hbtp]
    \centering
    \caption{Comparison with different diffusion steps.}
    \begin{tabular}{c|cc|cc}
    \hline
    \multirow{2}{*}{No.} & \multicolumn{2}{c|}{MSMT17}&  \multicolumn{2}{c}{DukeMTMC}\\
      &mAP &Rank-1 &mAP &Rank-1\\
    \hline
    Baseline   &75.4    &89.3    &82.8   &90.7\\
      2          &77.0    &90.2  &84.4  &92.1\\
      4          &77.5    &90.5  &84.6  &92.3\\
      6          &78.0    &90.9  &85.0  &92.0\\
      8          &78.6    &91.0  &85.3  &92.2\\
      10         &\textbf{79.0}    &{91.2}  &\textbf{85.6}  &\textbf{92.4}\\
      12         &{78.9}    &\textbf{91.3}  &{85.3}  &{92.1}\\
    \hline
    \end{tabular}
    \label{tab:time step}
\end{table}

\begin{table*}[htbp]
        \centering
        \caption{Effect of MEC with visual and guided features.}
        \resizebox{0.8\textwidth}{!}{
        \begin{tabular}{c l| c c| c c |c}
        \hline
        \multirow{2}{*}{MEC} &
        \multirow{2}{*}{Feature} &
        \multicolumn{2}{c|}{Performance} &
        \multicolumn{2}{c|}{Cosine Similarity} &
        \multirow{2}{*}{Generation Consistency $\uparrow$} \\
        \cline{3-6}
        & & mAP & Rank-1
        & Intra-class $\uparrow$ & Inter-class $\downarrow$ & \\
        \hline

        \multirow{2}{*}{w/o}
        & Visual
        & 77.0 & 89.5
        & 0.72 & 0.35
        & -- \\

        & Guided
        & 77.7 & 90.8
        & 0.76 & 0.29
        &0.75 \\
        \hline

        \multirow{2}{*}{w/}
        & Visual
        & \textbf{77.8} & \textbf{90.3}
        & \textbf{0.78} & \textbf{0.29}
        & -- \\

        & Guided
        & \textbf{79.0} & \textbf{91.2}
        & \textbf{0.83} & \textbf{0.24}
        & \textbf{0.89} \\
        \hline
        \end{tabular}
        \label{tab:mutual-enhancement effect}
        }
    \end{table*}
\begin{table}
    \centering
    \caption{Comparison with different guided features.}
    \begin{tabular}{c|cc|cc}
    \hline
    \multirow{2}{*}{No.}& \multicolumn{2}{c|}{MSMT17}&  \multicolumn{2}{c}{DukeMTMC}\\
              &mAP &Rank-1 &mAP &Rank-1\\
    \hline
      Baseline   &75.4    &89.3    &82.8   &90.7\\
      1          &76.7    &90.5    &84.1   &91.7\\
      2          &78.1    &90.9    &84.8   &92.0\\
      3          &\textbf{79.0}    &{91.2}    &\textbf{85.6}   &\textbf{92.4}\\
      4          &\textbf{79.0}    &\textbf{91.3}    &{85.4}   &{92.2}\\
    \hline
    \end{tabular}
    \label{tab:feature generation}
\end{table}

\textbf{Effect of the Total Number of Diffusion Steps.}
To analyze the effect of the total number of diffusion steps $T$, we conduct additional experiments in Tab.~\ref{tab:time step}.
As observed, increasing $T$ to 2 significantly boosts performance, reaching $77.0\%$ mAP on MSMT17 and $84.4\%$ mAP on DukeMTMC.
From $T=4$ to $T=8$, our model maintains competitive performance.
There is a trend that the discrimination of guided features gradually improves as $T$ increases.
With $T=10$, our model yields the best performance.
This result indicates that even a small total number of diffusion steps can enhance feature representation.

\textbf{Effect of Mutual Enhancement in MEC.}
In Tab.~\ref{tab:mutual-enhancement effect}, we show the effect of MEC with visual and guided features on MSMT17.
The MEC consistently improves the performance with both visual and guided features.
We further evaluate the intra-class and inter-class cosine similarities.
For each input image, three guided features are generated using independently sampled noises, and their average pairwise cosine similarity is used to measure the generation consistency.
With MEC, the intra-class similarity increases, while the inter-class similarity decreases.
The generation consistency also increases.
These results indicate that MEC produces more discriminative and stable guided features.
These results demonstrate that MEC benefits both features and support their mutual enhancement.

\textbf{Effect of Guided Features.}
In Tab.~\ref{tab:feature generation}, we evaluate the effect of guided features.
As observed, using one guided feature can improve the performance over the baseline.
This suggests that introducing guided features enhances the robustness of representation learning.
Increasing the number of guided features to two leads to further improvements, indicating that diverse guided features contribute to a more comprehensive feature.
When using three or four guided features, the model consistently achieves better performance.
To balance the performance and efficiency, we adopt three guided features as our default setting.

\textbf{Effect of Different Numbers of TCQ.}
In Tab.~\ref{tab:time step queries}, we evaluate the effect of using different numbers of TCQ.
As the number of TCQ increases from 2 to 16, the performance consistently improves on both MSMT17 and DukeMTMC.
This trend indicates that richer temporal conditions help the denoiser better model the evolving feature distribution across diffusion steps.
In particular, using 16 queries achieves the best results.
These results confirm the effectiveness of TCQ in enhancing the discrimination of guided features.
\begin{table}
\centering
\caption{Effect of different number of temporal queries.}
    \begin{tabular}{c|cc|cc}
    \hline
    \multirow{2}{*}{No.}& \multicolumn{2}{c|}{MSMT17}&  \multicolumn{2}{c}{DukeMTMC}\\
               &mAP &Rank-1 &mAP &Rank-1\\
    \hline
      2          &76.7    &89.9  &83.9  &91.4\\
      4          &77.1    &90.1  &84.1  &91.6\\
      8          &77.5    &90.6  &84.3  &92.1\\
      16         &\textbf{79.0}    &\textbf{91.2}  &\textbf{85.6}  &\textbf{92.4}\\
      32         &{78.7}    &{91.0}  &{85.4}  &\textbf{92.4}\\
    \hline
    \end{tabular}
    \label{tab:time step queries}
\end{table}

\textbf{Effect of Intermediate Features for the Generation.}
We investigate the impact of incorporating intermediate-layer features from the backbone as conditions for the denoiser.
In Tab.~\ref{tab:intermediate features}, $f_{12}$ refers to using only the output from the 12-\emph{th} layer as the condition, while $f_{12}+f_{11}$ indicates the addition of the outputs from the 11-\emph{th} and 12-\emph{th} layers as the condition.
These results demonstrate that features from earlier layers consistently improve the performance.
This suggests that intermediate-layer features provide additional information that enhances the generation.
Meanwhile, the results from the last line of Tab.~\ref{tab:intermediate features} also indicate that incorporating too many intermediate-layer features leads to a noticeable performance drop.
A plausible explanation is that early-layer features mainly capture low-level patterns and lack semantic discrimination, which may reduce the representational ability.
\begin{table}[]
\centering
\caption{Effect of intermediate features for the generation.}
    \resizebox{0.46\textwidth}{!}{
    \begin{tabular}{c|cc|cc}
    \hline
    \multirow{2}{*}{Setting}& \multicolumn{2}{c|}{MSMT17}&  \multicolumn{2}{c}{DukeMTMC}\\
      &mAP &Rank-1 &mAP &Rank-1\\
    \hline
      $f_{12}$          &77.8    &90.8  &84.4  &91.0\\
      $f_{12} + f_{11}$          &78.0    &91.2  &84.5  &91.2\\
      $f_{12} + f_{11} + f_{10}$          &78.6    &91.4  &84.6  &91.5\\
      $f_{12} + f_{11} + f_{10} + f_{9}$          &{78.8}    &{91.3}  &{84.9}  &{91.7}\\
      $f_{12} + f_{11} + f_{10} + f_{9} + f_{8}$          &{78.8}    &{91.0}  &{85.1}  &{92.0}\\
      $f_{12} + f_{11} + f_{10} + f_{9} + f_{8} + f_{7}$          &\textbf{79.0}    &\textbf{91.2}  &\textbf{85.6}  &\textbf{92.4}\\
      $f_{12} + f_{11} + f_{10} + f_{9} + f_{8} + f_{7} + f_{6}$          &{77.9}    &{90.8}  &{84.2}  &{91.8}\\
    \hline
    \end{tabular}}
    \label{tab:intermediate features}
\end{table}
\begin{table}[t]
    \centering
    \caption{Effect of different random seeds.}
    \label{tab:seed_robustness}
    \begin{tabular}{lc|cc|cc}
    \hline
    \multirow{2}{*}{Model} &\multirow{2}{*}{Seed} &\multicolumn{2}{c|}{MSMT17} &\multicolumn{2}{c}{DukeMTMC}\\
    &&mAP & Rank-1 &mAP & Rank-1 \\
    \hline
    Baseline & 1234 & 75.4 & 89.3 & 82.8 & 90.7 \\
    DiffReID & 1    & 78.8 & 91.3 & 85.8 & 92.3\\
    DiffReID & 12   & 78.4 & 91.4 & 85.5 & 92.0\\
    DiffReID & 123  & 78.7 & 90.7 & 85.4 & 92.2\\
    DiffReID & 1234 & 79.0 & 91.2 & 85.6 & 92.4\\
    \hline
    \end{tabular}
\end{table}
\begin{table}[t]
    \centering
    \caption{Effect of different diffusion features.}
    \label{tab:diffusion_target}
    \resizebox{0.48\textwidth}{!}{
    \begin{tabular}{l|cc|cc}
    \hline
    \multirow{2}{*}{Setting} &\multicolumn{2}{c|}{MSMT17} &\multicolumn{2}{c}{DukeMTMC}\\
    & mAP & Rank-1 & mAP & Rank-1 \\
    \hline
    Baseline & 75.4 & 89.3 & 82.8 & 90.7\\
    On visual features & 77.8 & 90.5 & 84.5 & 91.8\\
    On identity-aware text features & 79.0 & 91.2 & 85.6 & 92.4\\
    \hline
    \end{tabular}}
\end{table}
\begin{table}[t]
    \centering
    \caption{Effect of the KL regularization in VNG.}
    \label{tab:kl_ablation}
    \resizebox{0.45\textwidth}{!}{
    \begin{tabular}{c|cc|ccc}
    \hline
    \multirow{2}{*}{Model} &\multicolumn{2}{c}{MSMT17} &\multicolumn{2}{|c}{DukeMTMC}\\
    & mAP & Rank-1 & mAP & Rank-1 \\
    \hline
    Baseline & 75.4 & 89.3 & 82.8 & 90.7 \\
    DiffReID w/o KL  & 77.6 & 90.8 & 83.7 &91.1\\
    DiffReID  & 79.0 & 91.2 & 85.6 & 92.4\\
    \hline
    \end{tabular}}
\end{table}

\textbf{Effect of Different Random Seeds.}
To evaluate the training robustness, we conduct additional experiments with four different random seeds.
As shown in Tab.~\ref{tab:seed_robustness}, our method consistently outperforms the baseline under all random seeds.
Specifically, the mAP ranges from 78.4\% to 79.0\%, and the Rank-1 accuracy ranges from 90.7\% to 91.4\%.
The performance fluctuation is very small, indicating that our method is not sensitive to the random seed initialization.
These results demonstrate that the improvement of our method is stable and does not come from a specific random seed initialization.

\textbf{Effect of Different Diffusion Features.}
As shown in Tab.~\ref{tab:diffusion_target}, performing diffusion on visual features improves the baseline from 75.4\% to 77.8\% mAP and from 89.3\% to 90.5\% Rank-1 on MSMT17.
This indicates that diffusion-based feature distribution learning is beneficial for discriminative representation learning.
Furthermore, performing diffusion on identity-aware text features achieves better performance, reaching 79.0\% mAP and 91.2\% Rank-1.
These results demonstrate that identity-aware text features provide a more suitable semantic space for diffusion modeling.
As a result, the model can learn more discriminative and generalizable identity-aware feature distributions rather than appearance-biased visual feature distributions.

\textbf{Effect of the KL Regularization.}
As shown in Tab.~\ref{tab:kl_ablation}, removing the KL regularization still achieves 77.6\% mAP and 90.8\% Rank-1, outperforming the baseline by 2.2\% mAP and 1.5\% Rank-1.
This indicates that the visual-guided noise initialization in VNG is beneficial for discriminative feature generation.
With the KL regularization, the performance is further improved to 79.0\% mAP and 91.2\% Rank-1.
These results demonstrate that the KL regularization helps constrain the learned noise distribution and stabilizes the diffusion process, leading to more effective guided-feature generation.
\begin{table}[t]
\centering
\caption{Cost comparison with different methods.}
\label{tab:inference cost}
\renewcommand{\arraystretch}{1.15}
\resizebox{\columnwidth}{!}{
        \begin{tabular}{lccccc}
            \hline
            Method &
            \makecell[c]{Trainable Params\\(M)} &
            \makecell[c]{Memory\\(GB)} &
            \makecell[c]{GFLOPs\\} &
            \makecell[c]{Time\\(s/batch)}&
            \makecell[c]{mAP\\(MSMT17)} \\
            \hline
            SD-ReID~\cite{wang2026sd} & 103.36  & 11.05  & 677.67  & 1.66 &-- \\
            FusionReID~\cite{wang2025unity} & 153.80  & 1.62  & 28.10  & 0.14 &69.5 \\
            TF-CLIP~\cite{tfclip} &104.26  &1.53  &24.24  &0.26  &73.9\\
            DiffReID & 104.81 & 1.32  & 43.40  & 0.22 &79.0\\
            \hline
\end{tabular}
}
\end{table}
\subsection{Computational Cost Analysis}
Tab.~\ref{tab:inference cost} compares the computational cost of different methods.
Note that, our LWD adopts an MLP-based structure instead of the complex U-Net.
Thus, it is more efficient than previous diffusion-based methods, such as SD-ReID~\cite{wang2026sd}.
Compared with other methods, our method has a comparable computational overhead but achieves superior ReID performance.
These results demonstrate that our method provides a trade-off between computational cost and performance.
\begin{figure}[htbp]
    \centering
    \includegraphics[width=1.0\linewidth]{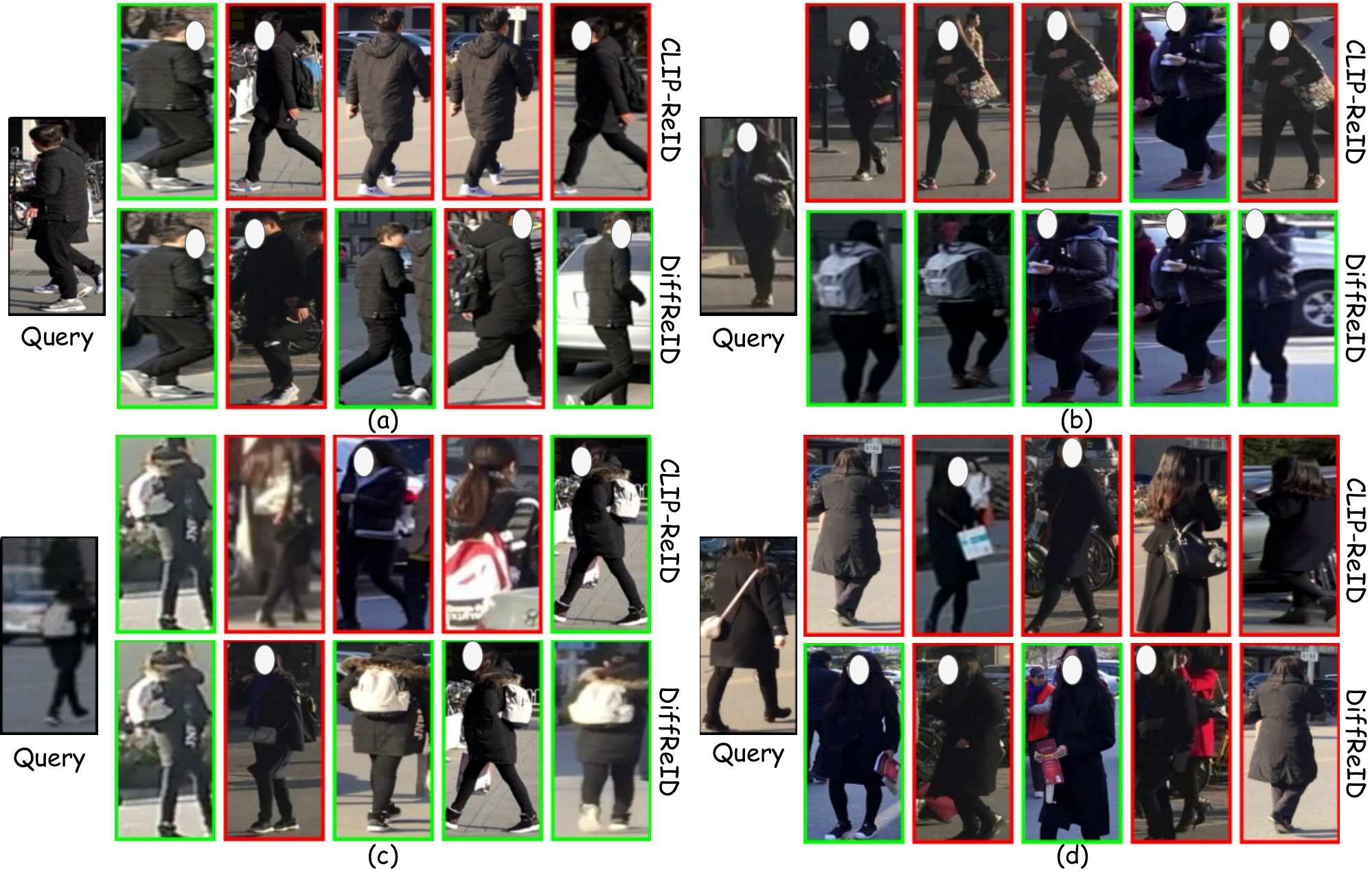}
    \caption{Retrieval results on MSMT17. Black, green and red boxes indicate query images, correct matches and incorrect matches, respectively.}
    \label{fig:rank}
\end{figure}
\begin{figure*}[htbp]
    \centering
    \includegraphics[width=1.0\linewidth]{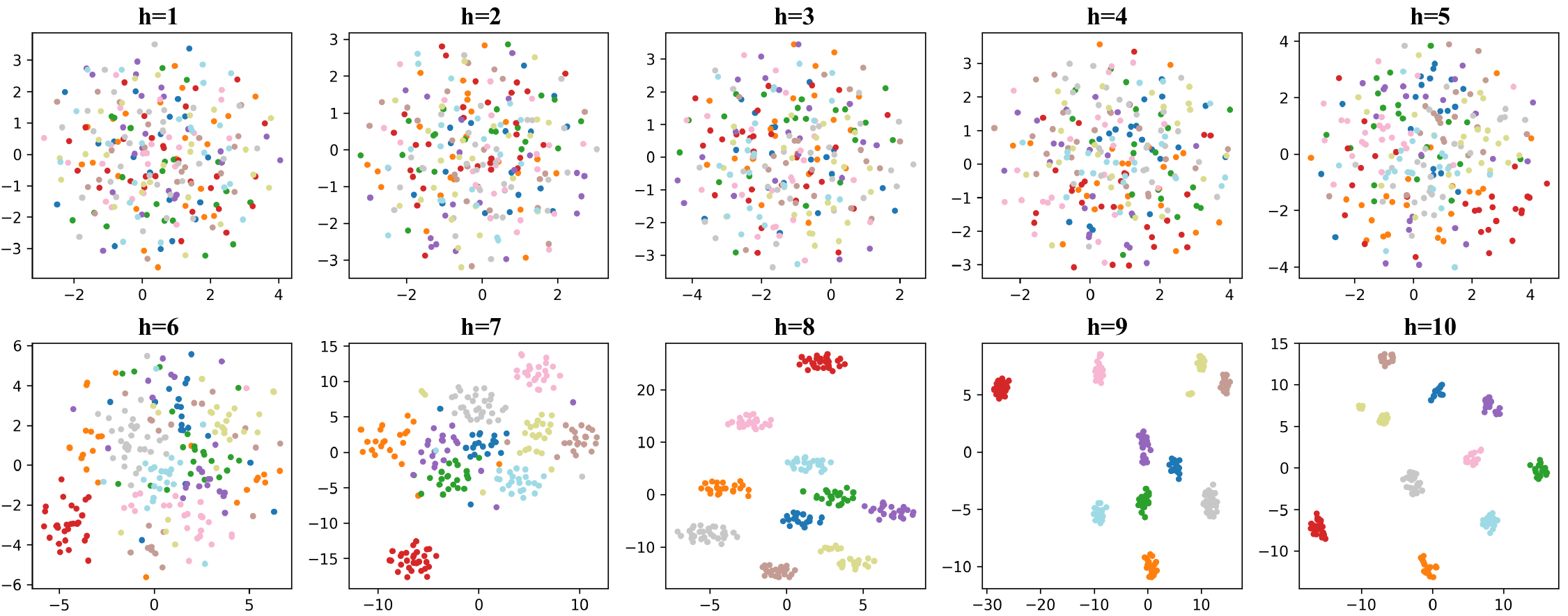}
    \caption{t-SNE visualization of guided features at successive denoising timesteps.}
    \label{fig:diffusion_process}
\end{figure*}
\subsection{Qualitative Analysis}
To better understand our proposed framework, we present comprehensive qualitative results in this subsection.

\textbf{Top-5 Retrieval Results.}
Fig.~\ref{fig:rank} shows retrieval results with CLIP-ReID~\cite{clip_reid} and our method.
In general, our method captures more fine-grained information while reducing the influence of local salient features.
In Fig.~\ref{fig:rank}(a) and (b), CLIP-ReID is distracted by visually prominent attributes, such as similar black clothing and white backpack, leading to incorrect matches.
In contrast, our method focuses on more comprehensive information.
It avoids over-reliance on a few specific attributes and successfully identifies the correct match among visually similar samples.
A similar pattern is observed in Fig.~\ref{fig:rank}(c) and (d).
In Fig.~\ref{fig:rank}(d), even though the person in the query image carries a white backpack, our method does not get misled by this prominent detail.
In Fig.~\ref{fig:rank}(c), despite significant distortions in the query image, our method still accurately retrieves the correct match.
\begin{figure}[htbp]
    \centering
    \includegraphics[width=1.0\linewidth]{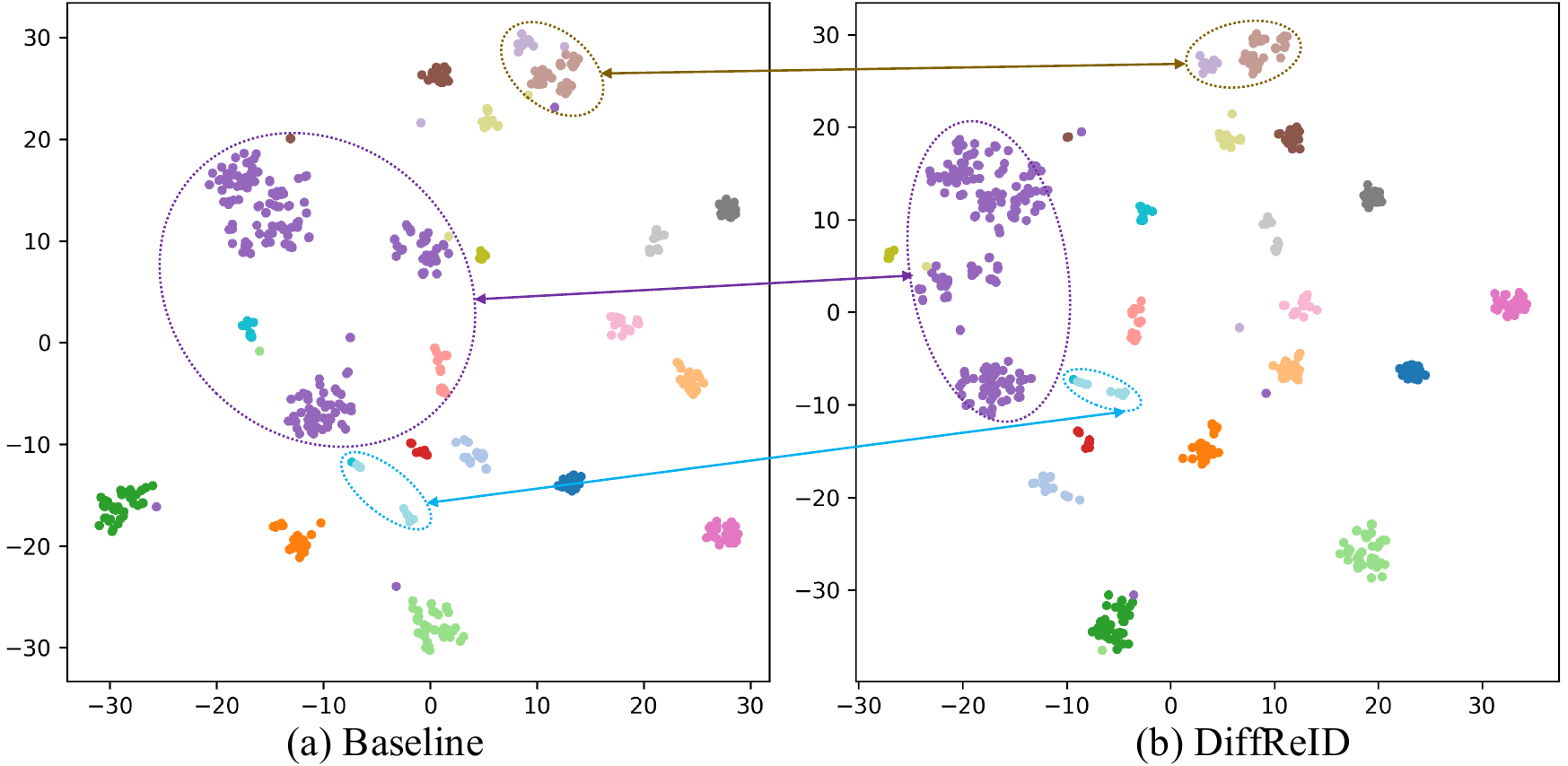}
    \caption{Feature distributions visualized by t-SNE. Different colors represent different identities.}
    \label{fig:tsne}
\end{figure}
\begin{figure}[htbp]
    \centering
    \includegraphics[width=1.0\linewidth]{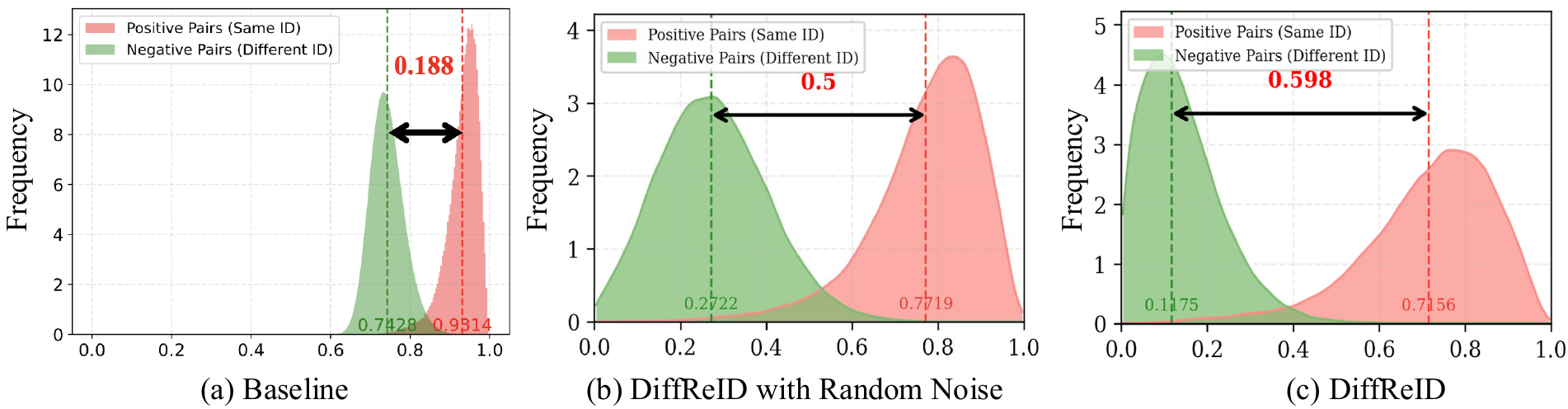}
    \caption{Visualization of the cosine similarity distribution.}
    \label{fig:inter-intra_dist}
\end{figure}

\textbf{Visualization of Feature Distribution.}
We further analyze the feature distributions on the MSMT17 datasets.
Specifically, as shown in Fig.~\ref{fig:tsne}, we randomly sample twenty identities from the MSMT17 test set and visualize their feature distributions using t-SNE~\cite{tsne}.
Each point represents a sample and different colors represent different identities.
Compared with the baseline, our DiffReID can better reduce intra-class distances and increase inter-class distances.
The main reason is that our DiffReID directly captures the identity-aware distribution rather than the salient semantic regions.

Furthermore, we visualize the cosine similarity distributions on the MSMT17 test set to analyze the discrimination of different feature representations.
As shown in Fig.~\ref{fig:inter-intra_dist}(a), the baseline exhibits a small distribution gap between positive and negative pairs.
It indicates that the baseline struggles to distinguish samples with similar visual patterns.
When our DiffReID is introduced with random Gaussian noises, as shown in Fig.~\ref{fig:inter-intra_dist}(b), the distribution gap becomes much larger.
It demonstrates that diffusion-based guided feature generation improves feature discrimination.
Moreover, when visual-guided noises are adopted in Fig.~\ref{fig:inter-intra_dist}(c), the negative-pair distribution further shifts to a lower similarity range while the positive-pair distribution remains in a high-similarity range.
As a result, the distribution gap increases from 0.5 to 0.598.
This indicates that VNG provides a more suitable noise initialization for generating discriminative features.
It improves the inter-identity separation while preserving the intra-identity consistency.
Overall, these results show that our DiffReID captures identity-aware distributions and obtains more discriminative representations.

\textbf{Visualization of Guided Features Across the Diffusion Process.}
To investigate how the model gradually captures the identity-aware feature distribution, we visualize guided features at successive denoising timesteps using t-SNE, as shown in Fig.~\ref{fig:diffusion_process}.
Specifically, we randomly sample ten identities from the MSMT17 test set.
Each point represents a sample and different colors denote different identities.
At the early denoising timesteps, the guided features exhibit large variance and strong overlap across identities, indicating that they primarily encode coarse and noisy structural information.
As the diffusion denoising process progresses, the features begin to form clearer local structures, and samples belonging to the same identity become increasingly aggregated.
In the later denoising timesteps, the guided features evolve into well-separated and compact clusters for each identity.
It demonstrates that the denoiser effectively captures discriminative semantic cues and refines the representation toward the underlying identity-aware distribution.
These observations confirm that our model progressively improves the discrimination of guided features through step-wise denoising.
Noted that, the feature shift between $h=6$ and $h=7$ arises from the frequency-dependent reconstruction behavior in the diffusion model~\cite{yi2024working}.
Specifically, early denoising steps mainly recover low-frequency components that encode global structures.
Later denoising steps progressively restore high-frequency details, introducing fine-grained identity cues.
Thus, it reflects the shift from low-frequency structure to high-frequency identity, rather than instability in the denoising process.
\begin{figure}[htbp]
    \centering
    \includegraphics[width=1.0\linewidth]{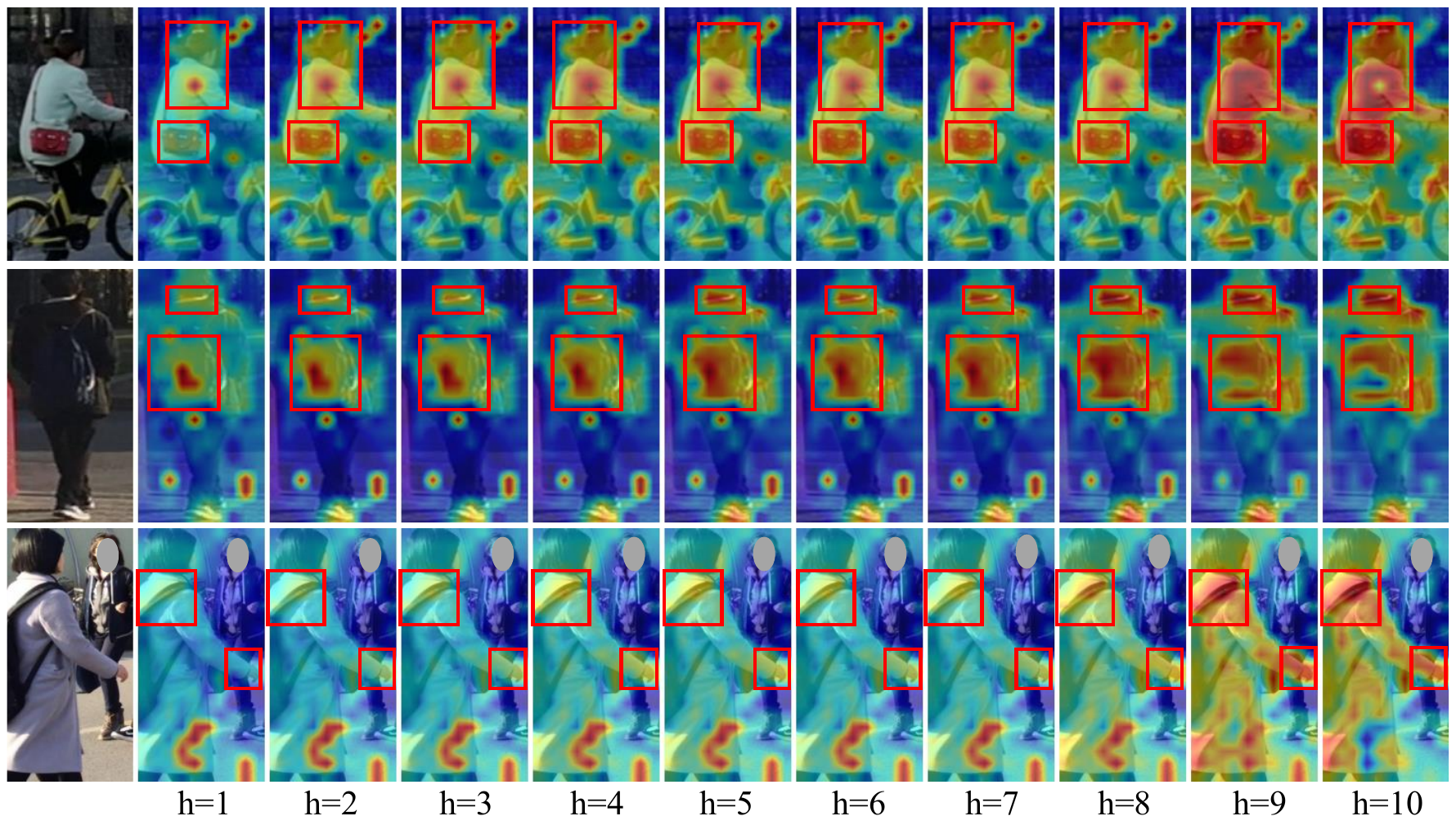}
    \caption{Visualization of attention maps for different temporal queries. Darker red indicates higher attention weights.}
    \label{fig:cond_vis}
\end{figure}

\textbf{Visual Effect of TCQ.}
To better understand the effect of TCQ, we visualize attention maps of learned queries at different timesteps.
As shown in Fig.~\ref{fig:cond_vis}, our method attends to different information at different timesteps.
In early denoising steps, it focuses more on the global structure of persons.
In later denoising steps, it gradually shifts attention to discriminative regions highlighted by the red boxes.
Moreover, as shown in Fig.\ref{fig:cond_vis}, our method can effectively suppress irrelevant distractions.
Notably, in the last row of Fig.~\ref{fig:cond_vis}, even when an irrelevant person appears, our method still focuses on the correct target.
In summary, these visualization results indicate that our method progressively generates discriminative features while avoiding the influence of irrelevant contents.
\section{Conclusion}
In this paper, we propose DiffReID, an advanced generative framework for image-based object ReID.
It learns identity-aware distributions and generates diverse guided features to complement visual features for robust representation.
The framework consists of a Vision-guided Noise Generator (VNG) to sample visual-guided noises and a Light Weight Denoiser (LWD) to model identity-aware distributions.
In addition, a Mutual Enhancement Constraint (MEC) is introduced to facilitate mutual learning between visual features and guided features, thereby improving both generalization and discrimination.
Extensive experiments validate that our proposed method performs better than existing works on several widely used object ReID benchmarks.
\bibliographystyle{IEEEtran}
\bibliography{IEEEabrv,main}
\end{document}